\documentclass[dvipsnames]{valence} %

\usepackage{soul} %
\usepackage{float}
\usepackage{amsmath}
\usepackage{algorithm}
\usepackage{algpseudocode}
\usepackage{amsfonts}
\usepackage{subcaption}
\usepackage{longtable}

\usepackage[export]{adjustbox}
\usepackage{comment}
\usepackage{mathtools}
\usepackage{authblk}
\usepackage{fancyhdr}
\usepackage{multirow}
\usepackage{booktabs}
\usepackage{array}
\usepackage{hyperref}
\usepackage{cleveref}

\usepackage{amsmath,amsfonts,bm}

\def\eqref#1{equation~\ref{#1}}

\def\1{\bm{1}}

\DeclareMathAlphabet{\mathsfit}{\encodingdefault}{\sfdefault}{m}{sl}
\SetMathAlphabet{\mathsfit}{bold}{\encodingdefault}{\sfdefault}{bx}{n}

\newcounter{boxcounter}
\crefname{boxcounter}{Box}{Boxes}

\title{Virtual Cells: Predict, Explain, Discover}

\setrunningtitle{Virtual Cells: Predict, Explain, Discover}

\author[\star]{Emmanuel Noutahi}
\author[\star]{Jason Hartford}
\author{Prudencio Tossou}
\author{Shawn Whitfield}
\author{Alisandra K. Denton}
\author{Cas Wognum}
\author{Kristina Ulicna}
\author{Michael Craig}
\author{Jonathan Hsu}
\author{Michael Cuccarese}
\author{Emmanuel Bengio}
\author{Dominique Beaini}
\author{Christopher Gibson}
\author{Daniel Cohen}
\author[\star]{Berton Earnshaw}

\affiliation{Valence Labs, Recursion}
\contribution[\star]{\small Correspondence: \{firstname\}@valencelabs.com}

\abstract{
Drug discovery is fundamentally a process of inferring the effects of treatments on patients, and would therefore benefit immensely from computational models that can reliably simulate patient responses, enabling researchers to generate and test large numbers of therapeutic hypotheses safely and economically before initiating costly clinical trials. Even a more specific model that predicts the functional response of cells to a wide range of perturbations would be tremendously valuable for discovering safe and effective treatments that successfully translate to the clinic. Creating such \textit{virtual cells} has long been a goal of the computational research community that unfortunately remains unachieved given the daunting complexity and scale of cellular biology. Nevertheless, recent advances in AI, computing power, lab automation, and high-throughput cellular profiling provide new opportunities for reaching this goal. In this perspective, we present a vision for developing and evaluating virtual cells that builds on our experience at Recursion. We argue that in order to be a useful tool to \textit{discover} novel biology, virtual cells must accurately \textit{predict} the functional response of a cell to perturbations and \textit{explain} how the predicted response is a consequence of modifications to key biomolecular interactions. We then introduce key principles for designing therapeutically-relevant virtual cells, describe a lab-in-the-loop approach for generating novel insights with them, and advocate for biologically-grounded benchmarks to guide virtual cell development. Finally, we make the case that our approach to virtual cells provides a useful framework for building other models at higher levels of organization, including virtual patients. We hope that these directions prove useful to the research community in developing virtual models optimized for positive impact on drug discovery outcomes. 
}

\begin{document}

\maketitle

\section{Introduction}\label{sec:intro}

Drug discovery is fundamentally a process of accurately inferring the effects of treatments on patients. Unfortunately, it is notoriously costly and riddled with failure \citep{wong2019estimation,jones2018biomedical,dimasi2016innovation,paul2010improve}. Despite decades of innovations, for every ten drugs that enter clinical trials today, roughly nine of those will fail to receive approval, representing unfortunate delays in addressing patient needs, substantial losses in R\&D investment, and a significant deficit in our collective understanding of human physiology and pathology. Nevertheless, the impact of each approved therapy on the lives of patients, particularly those addressing unmet need, is hard to overstate, thus any approach that meaningfully improves our ability to correctly predict the effect of treatments in patients would be of immense value to both patients and drug discoverers alike.

One such approach is the computational simulation of therapeutic interventions in \textit{virtual patients}, or mechanistic models accounting for the physiological factors necessary to accurately infer patient-level response to treatments. Virtual patients could revolutionize drug discovery by enabling researchers to generate and test large numbers of therapeutic hypotheses safely and economically before initiating costly clinical trials. However, though simulation has already revolutionized a number of industries \citep{winsberg2019science,singh2022applications}, examples of practical and effective simulation in drug discovery are rare due to the challenges inherent in modeling the scale and complexity of biological systems \citep{ideker2001integrated,goldberg2018emerging,georgouli2023multi}. Even the simulation of a single prokaryotic cell is daunting \citep{karr2012whole}, and simulating the full complexity of a eukaryotic cell lies beyond current capabilities \citep{georgouli2023multi}. Nevertheless, the ability to faithfully simulate the effect of therapeutic interventions at any level of biological organization---cell, tissue, organ, patient---in a corresponding \textit{virtual model} has the potential to significantly improve drug discovery outcomes.

\subsection{The Predict-Explain-Discover capabilities of virtual models}\label{sec:intro-ped}

What exactly makes virtual models so potentially valuable for drug discovery? They offer the ability to accurately:

\begin{enumerate}
    \item \textit{predict} the effects of interventions on the model system,
    \item \textit{explain} the predicted response in terms of one or more changes to supporting mechanisms, and
    \item \textit{discover} novel insights by generating and testing therapeutic hypotheses.
\end{enumerate}

To better understand what we mean, we give two examples using well-known cancer treatments; these are only meant to motivate the concepts and not to imply that these would represent novel discoveries if made today:

\textbf{Vorinostat.} A virtual cell could \textit{predict} that treatment with Vorinostat up-regulates tumor suppressor genes, and \textit{explain} this effect by identifying inhibition of histone deacetylase (HDAC) enzymes as the underlying mechanism. Based on this understanding, the model could then \textit{discover} biomarkers of response,  or synergistic combinations such as with DNA-damaging agents like cisplatin, by revealing how chromatin decondensation increases sensitivity to genotoxic stress.

\textbf{Pembrolizumab.} For a virtual \textit{patient} model that accounts for mechanisms across different cells, tissues and organs, it could \textit{predict} that Pembrolizumab reduces tumor burden in various cancers, and \textit{explain} this effect by simulating how the drug blocks the Programmed Cell Death Protein 1 (PD-1) immune checkpoint to restore T-cell activity. Building on this understanding, the model could then \textit{discover} combination strategies to enhance efficacy across patient subgroups, as well as approaches to overcome emerging resistance mechanisms.

\begin{figure}[tb]
    \centering
    \includegraphics[width=0.9\textwidth, valign=t]{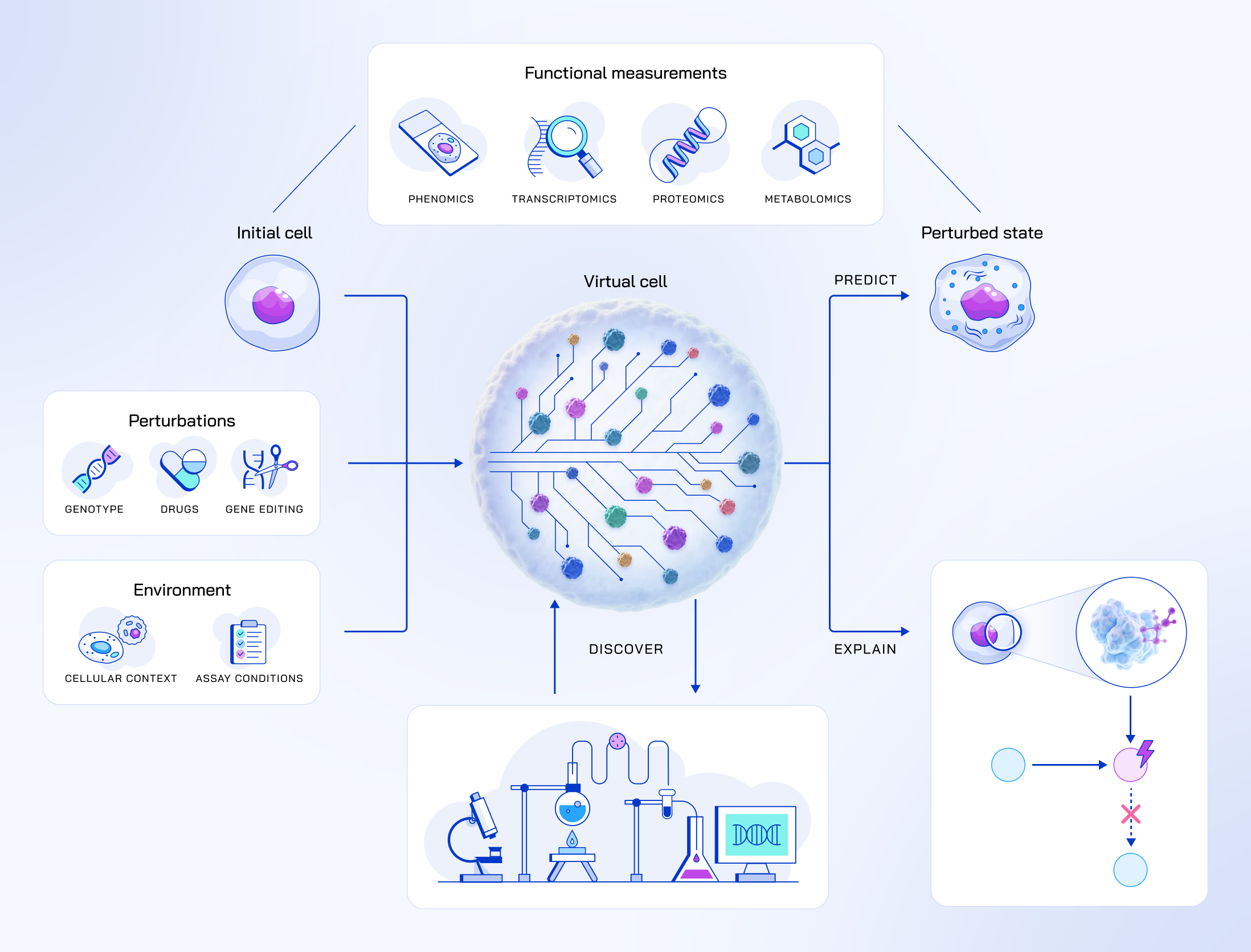}
    \caption{\small
    \textbf{Virtual cells: predict, explain, discover.} In order to address critical issues holding back progress in drug discovery, virtual cells should \textit{predict} the functional response of cells (as measured in phenomics, transcriptomics, proteomics, metabolomics, etc.)~to perturbations across diverse cellular contexts and assay conditions, and \textit{explain} these responses as modifications to key biomolecular interactions, using techniques like causal learning on interventional data, physics-informed structural predictions, and targeted molecular dynamics simulations. By offering a focused mechanistic understanding of their behavior, such virtual cells can \textit{discover} novel biology by efficiently generating and testing large numbers of promising hypotheses before initiating costly clinical trials, offering a modern approach to rational drug design that holistically optimizes cell state rather than biomolecular interactions only.
    }\label{fig:vc}
\end{figure}

Throughout this paper, we will refer to these three capabilities as the \textit{Predict-Explain-Discover}, or \textit{P-E-D}, capabilities of virtual models, and claim that it is precisely the ability of virtual models to accurately \textit{predict} outcomes and \textit{explain} them mechanistically that would make them powerful tools to \textit{discover} novel therapeutic insights. As our prototypical example of a virtual model, we illustrate the P-E-D capabilities for virtual cells in \Cref{fig:vc}  and describe these capabilities in more detail in Box~\ref{box:vc-capabilities}.

\refstepcounter{boxcounter}
\begin{tcolorbox}[title={Box~\theboxcounter: Virtual Cells: Predict, Explain, Discover}, colback=valbg,float*=t]
\label{box:vc-capabilities}

\textbf{Predict} the functional response of cells to perturbations across diverse biological contexts, timepoints and modalities. This includes modeling gene expression, morphology, protein activity, and other phenotypic changes under genetic or chemical interventions.

\vspace{0.5em}

\textbf{Explain} these responses by identifying key biomolecular interactions, causal pathways, and context-dependent regulatory mechanisms. Correct explanations support predictions by enabling both generalization beyond the training data and reasoning about counterfactuals and the response of biological systems at higher levels of organization.

\vspace{0.5em}

\textbf{Discover} new biological insights and actionable therapeutic hypotheses through lab-in-the-loop experimentation, using the virtual cell as a world model for systematic hypothesis generation, testing, and refinement.

\end{tcolorbox}

We wish to further clarify what we mean by these terms. By \textit{predict} we refer to the task of a virtual model to predict the effect of a perturbation on a biological system, typically a cell but applicable at all levels of organization (see \Cref{sec:intro-ped}).  In causal terms, this is equivalent to estimating (the distribution of) outcomes under intervention \citep{pearl2009causality}. 
We use \textit{explain} in a broader, biologically-grounded sense than is common in machine learning, where explainability usually refers to post-hoc techniques (e.g., saliency maps, attention scores, feature attributions) used to interpret model behavior. In contrast, \textit{explanation} in our context refers to structured, testable accounts of how a given perturbation leads to a specific biological response. These accounts must be meaningful in the language of biology and useful for hypothesis generation and falsification. See Box~\ref{box:explain} for more on biological explanations.

\begin{figure}[t]
\input{boxes/explain}
\end{figure}

\subsection{Virtual models without fully mechanistic simulation}\label{sec:intro-advances}

Given the current difficulties building fully mechanistic virtual model simulators, the question naturally arises: can we build these models without resorting to fully mechanistic simulation? We argue that four recent advances put this objective within reach today, particularly for virtual cells:

\begin{enumerate}
    \item modern AI and machine learning (\textbf{AI/ML}),
    \item modern \textbf{compute infrastructure},
    \item \textbf{automated labs} for high-throughput cellular data generation, and
    \item the proliferation of \textbf{cellular omics datasets}.
\end{enumerate}

We briefly describe each of these advances below:

\paragraph{AI/ML.} While traditional computational drug discovery techniques have struggled to deal with biological complexity \citep{sams2005target,swinney2011were,waring2015analysis}, recent advancements in AI/ML have enabled the training of powerful models capable of extracting patterns from high-dimensional biological datasets and predicting complex biological phenomena \citep{zhang2025artificial,wang2023scientific,sadybekov2023computational,vamathevan2019applications}. Modern AI/ML also provides techniques for integrating multimodal measurements that are essential for capturing a complete view of cells \citep{nam2024harnessing, stahlschmidt2022multimodal,li2018review,picard2021integration}. Of course, scientific progress depends not only on accurate predictions but on understanding the mechanisms that give rise to these predictions, and here we can turn to modern AI/ML again for methods that leverage interventional datasets to train models that directly predict causal effects and infer causal mechanisms \citep{scholkopf2021toward,pearl2009causality,hill2016inferring,sachs2005causal}. 

\paragraph{Compute infrastructure.} Recent improvements in the capability and accessibility of computational infrastructure, driven by dedicated on-premise installations, scalable cloud platforms, specialized hardware accelerators (e.g., GPUs, TPUs), and high-bandwidth networking, supports the training of powerful AI/ML models on high-dimensional biological datasets \citep{lee2018exascale,zhou2024training}.

\paragraph{Automated labs.} Advances in automation technology, including robotics for plate and liquid handling and high-throughput microscopy, make possible the building of sophisticated automated labs that generate biological data at the scale, quality, and diversity required for training useful AI/ML models. For example, the automated phenomics lab we operate at Recursion alone is capable of obtaining microscopy readouts from $\sim$2.2 million samples per week.

\paragraph{Cellular omics datasets.} We have recently witnessed a rapid expansion in the availability of both public and private cellular omics datasets---genomic, transcriptomic, proteomic, metabolomic, phenomic---as well as improved techniques for data integration and curation, providing the raw materials necessary for training powerful AI/ML models \citep{Fay2023, gkae1142, Zhang2025, Chandrasekaran2024}.

Drawing on more than a decade of experience at Recursion building predictive, generative, and causal models to accelerate drug discovery, we believe that we can build virtual models with P-E-D capabilities without resorting to fully mechanistic simulation by training AI/ML models on massive interventional datasets using powerful compute infrastructure combined with active lab-in-the-loop data generation. In so doing, we will realize the transformative impact we believe virtual models will have on drug discovery (see \Cref{tab:vc-for-dd} for examples of how virtual cells could have this impact).

\begin{table}[htbp]

\centering

\caption{\textbf{Applications of virtual cells in drug discovery.} Virtual models at every level of biological organization could revolutionize the drug discovery process, from early disease modeling through clinical trial design, by aiding researchers to \emph{predict} response to therapies, \emph{explain} response via key mechanisms, and \emph{discover} novel insights through lab-in-the-loop experimentation. Here we outline how the Predict-Explain-Discover capabilities of virtual cells could have this impact.}
\label{tab:vc-for-dd}

\footnotesize

\renewcommand{\arraystretch}{1.2}

\begin{tabular}{m{3.1cm} m{8.2cm} m{3.0cm}}

\toprule

\textbf{Drug discovery stage} & \textbf{Applications} & \textbf{Capabilities} \\

\midrule

\multirow[c]{4}{=}{Understanding Disease Mechanisms} 
& Compare healthy vs.~diseased states to identify perturbed regulatory mechanisms and disease-specific vulnerabilities 
& Explain, Discover \\ \cline{2-3}
& Explain how genetic backgrounds alter disease mechanisms, variability in disease manifestation, and drug responses to identify robust, context-specific druggable entry points 
& Explain \\ \hline

\multirow{5}{=}{Target Identification \& \\Validation} 
& Discover and prioritize disease-driving genes by simulating the functional consequences of mutations, loss-of-function events, splicing variants, and dysregulated expression 
& Explain, Discover \\ \cline{2-3}
& Predict target essentiality (pan-cell or context-specific) and co-dependencies (e.g., synthetic lethality) 
& Predict \\ \cline{2-3}
& Predict target druggability and downstream effects of modulating a specific target in disease-relevant contexts 
& Predict \\ \hline

\multirow{3}{=}{Hit Identification \& \\Compound Screening} 
& Perform large-scale virtual screens of compounds, predicting activity across multiple cell lines and contexts 
& Predict \\ \cline{2-3}
& Predict compound selectivity and off-target effects across cell types (e.g., toxicity versus efficacy) 
& Predict \\ \hline

\multirow{6}{=}{Mechanism of Action Studies} 
& Map compound phenotypic responses to upstream molecular events and generate plausible MoA hypotheses through reasoning over structural and functional data 
& Explain, Discover \\ \cline{2-3}
& Explain polypharmacology using multimodal perturbation signatures 
& Explain \\ \cline{2-3}
& Predict molecular and phenotypic outcomes following compound perturbation, capturing both acute (short-term) and chronic (long-term) response dynamics 
& Predict \\ \hline

\multirow{5}{=}{Hit-to-Lead \& Lead Optimization} 
& Predict and explain structure-activity relationships (SAR) to guide minimal structural modifications that enhance efficacy, optimize selectivity, or reduce liabilities 
& Predict, Explain \\ \cline{2-3}
& Predict ADMET profiles to optimize pharmacokinetic and safety properties  
& Predict \\ \cline{2-3}
& Identify mechanisms and guide designs for emerging therapeutic modalities (allosteric modulators, covalent inhibitors, and glues) 
& Explain, Discover \\ \hline

\multirow{5}{=}{Resistance Prediction \& Disease Evolution} 
& Predict and explain emergence of drug resistance through pathway rewiring, feedback loops, or network-level adaptation 
& Predict, Explain \\ \cline{2-3}
& Predict clonal evolution dynamics and selection pressures in response to therapeutic interventions 
& Predict \\ \cline{2-3}
& Discover rational combination therapies or synthetic lethality strategies to overcome or delay resistance 
& Discover \\ \hline

\multirow{5}{=}{Preclinical \& Translational Modeling} 
& Explain context-specific compound activity (e.g., toxicity in one tissue versus efficacy in another) 
& Explain \\ \cline{2-3}
& Predict therapeutic, immune, and inflammatory responses across patient-derived and experimental models  
& Predict \\ \cline{2-3}
& Discover robust biomarkers predictive of patient-specific therapeutic responses 
& Discover \\ \hline

\multirow{3}{=}{Clinical Trial Design \& Biomarker Strategy} 
& Inform patient stratification strategies and biomarker-based inclusion criteria 
& Discover \\ \cline{2-3}
& Predict optimal human dose and combination schedules for clinical studies 
& Predict \\

\bottomrule

\end{tabular}
\end{table}

In this perspective we share our vision of how to leverage these advances to build therapeutically-relevant virtual models. Here we focus on \textit{virtual cells}, due to both the heightened interest in these models currently, as well as the wide availability of large cellular datasets. We introduce key design principles for virtual cells, and describe a lab-in-the-loop paradigm for continuously refining virtual cells, treating them as testable theories of human cellular physiology and pathology that agentic systems attempt to falsify. We also advocate for biologically-meaningful benchmarks to guide virtual cell development, and make the case that the framework presented here is appropriate for building virtual models with Predict-Explain-Discover capabilities at all levels of biological organization. 

\subsection{Prior work and current perspectives on virtual cells}\label{sec:intro-prior-work}

The idea of building virtual models, especially virtual cells, is not new; in fact, virtual cells have been recognized as one of this century's ``grand challenges'' in computational biology \citep{tomita2001whole}. Early pioneering efforts such as E-Cell~\citep{tomita1999cell,tomita2001whole} and Virtual Cell (VCell)~\citep{loew2001virtual,slepchenko2003quantitative} created rule-based or reaction-diffusion frameworks that integrated diverse biological datasets to model cellular processes. These approaches were later extended to genome-scale whole-cell models for bacteria \citep{karr2012whole,macklin2020simultaneous,sun2021coli} and yeast \citep{ye2020comprehensive} which attempted to model a significant portion of genes, gene products, and their functions, demonstrating that it is, in principle, possible to predict phenotypes from genotypes. These models have been used to validate experimental findings, predict novel mechanisms, and enable counterfactual simulations. However, they remain limited to simple organisms and face major challenges in scalability, dynamic complexity, and computational tractability, and their construction remains labor-intensive, with only a handful built to date \citep{georgouli2023multi}. Importantly, these efforts have exposed a persistent gap between available biological data and the requirements for parameterizing large-scale mechanistic simulations. More recently, the field has also moved toward structural modeling approaches, including the reconstruction of whole-cell 3D structures, such as the bacterium \textit{M. Genitalium} \citep{maritan2022building}, and attempts to simulate the entire minimal cell JCVI-syn3A via coarse-grained molecular dynamics, although current MD engines were unable to complete the simulation due to computational limitations \citep{stevens2023molecular}.

Although not virtual cells per se, several large-scale initiatives have sought to advance our understanding of human biology by mapping the relationships among biological components. Examples include the  IUPS Human Physiome Project, which is developing a multi-scale framework for the hierarchical modeling of physiological function \citep{hunter2002iups,hunter2024physiome,hunter2003integration}, the Human Cell Atlas, which is producing comprehensive reference maps for all human cells \citep{regev2018human,rood2025human}, the Human Proteome Project, which aims to map all expressed proteins in the human body \citep{hanash2002human,Omenn_2024}, and the Human Connectome Project, which aims to map all neural connections in the human brain \citep{van2013wu}.

We note that several new perspectives have recently proposed updated visions for virtual cells in the era of AI and large-scale biological data. A group led by the Allen Institute for Cell Science advocates for the integration of top-down phenomenological models of cell behavior with bottom-up structural models of biomolecular interactions, using knowledge graphs to connect the various spatial and temporal scales \citep{johnson2023building}. Separately, a group sponsored by the Chan-Zuckerberg Initiative envisions virtual cells as collections of embeddings, generated by specific foundation models, of the various biomolecules found in cells, with ``virtual instruments'' designed to simulate interventions and predict associated readouts \citep{bunne2024build}. Similarly, a recent perspective \citep{cui2025towards} proposes building multimodal foundation models that integrate omics data across modalities via unified transformer architectures, enabling applications such as \textit{in silico} perturbation, cell state characterization, and biomarker discovery.

We share several points of agreement with these perspectives, notably the need to combine top-down and bottom-up modeling approaches, the critical role of multimodal data integration, and the transformative potential of foundation model architectures. However, our view on virtual cells is distinct in critical ways: while others emphasize static representations or predictive embeddings, we prioritize building causal, mechanistically-grounded models that not only predict but also explain the functional response of cells to perturbations. Given our primary objective of bringing new medicines to patients, we view virtual cells not simply as descriptive models, but as systems capable of iteratively generating interpretable and testable hypotheses of cellular behavior, which can be continuously refined through experimental feedback to drive therapeutic discovery.

\section{A Vision for Virtual Cells}\label{sec:vision}

Cell biology can be characterized at multiple scales. At the \textit{molecular level}, physical equations describe how forces govern the behavior of individual atoms. In principle, atomistic simulations that integrate the effects of these forces over time would allow for the exact simulation of an entire cell. However, doing so for the roughly 100 trillion atoms estimated to make up a typical eukaryotic cell~\citep{milo2015cell} remains computationally intractable. Still, the cumulative behavior of these trillions of atoms in response to an external perturbation, such as treatment with a drug, can be measured at the \textit{cellular level} via omics experiments. If these cellular-level measurements are sufficiently detailed, we then obtain a holistic view of the \textit{functional response} of the cell to the perturbation. 

For drug discovery, modeling system-wide cellular response is essential, since most therapeutic interventions do not act on a single isolated target but instead modulate networks of biomolecules, triggering feedback loops and producing off-target effects. Simply predicting the impact of a perturbation on a single protein or gene is insufficient for capturing this holistic, functional view. We therefore argue that virtual cells must first act as predictors of the functional response of cells to perturbations in order to be relevant for drug discovery---the first of the P-E-D capabilities.

Several recent efforts have demonstrated progress in this direction, proposing promising approaches to predicting transcriptomic and phenotypic changes following genetic or chemical perturbations, including GEARS~\citep{Roohani_2023}, CPA~\citep{Lotfollahi_2023}, scLAMBDA~\citep{wang2024modeling}, CellFlow~\citep{klein2025cellflow} and TxPert~\citep{wenkel2025txpert}. These models leverage information from the molecular scale to better predict functional response at the cellular scale, by using either embeddings of small molecules \citep{sypetkowski2024molgps} or proteins \citep{Hayes2024}, or biological knowledge such as protein-protein or pathway-based interaction networks \citep{Roohani_2023,wang2024modeling,wenkel2025txpert}.  

\begin{figure}[tb]
    \centering
    \includegraphics[width=0.9\textwidth, valign=t]{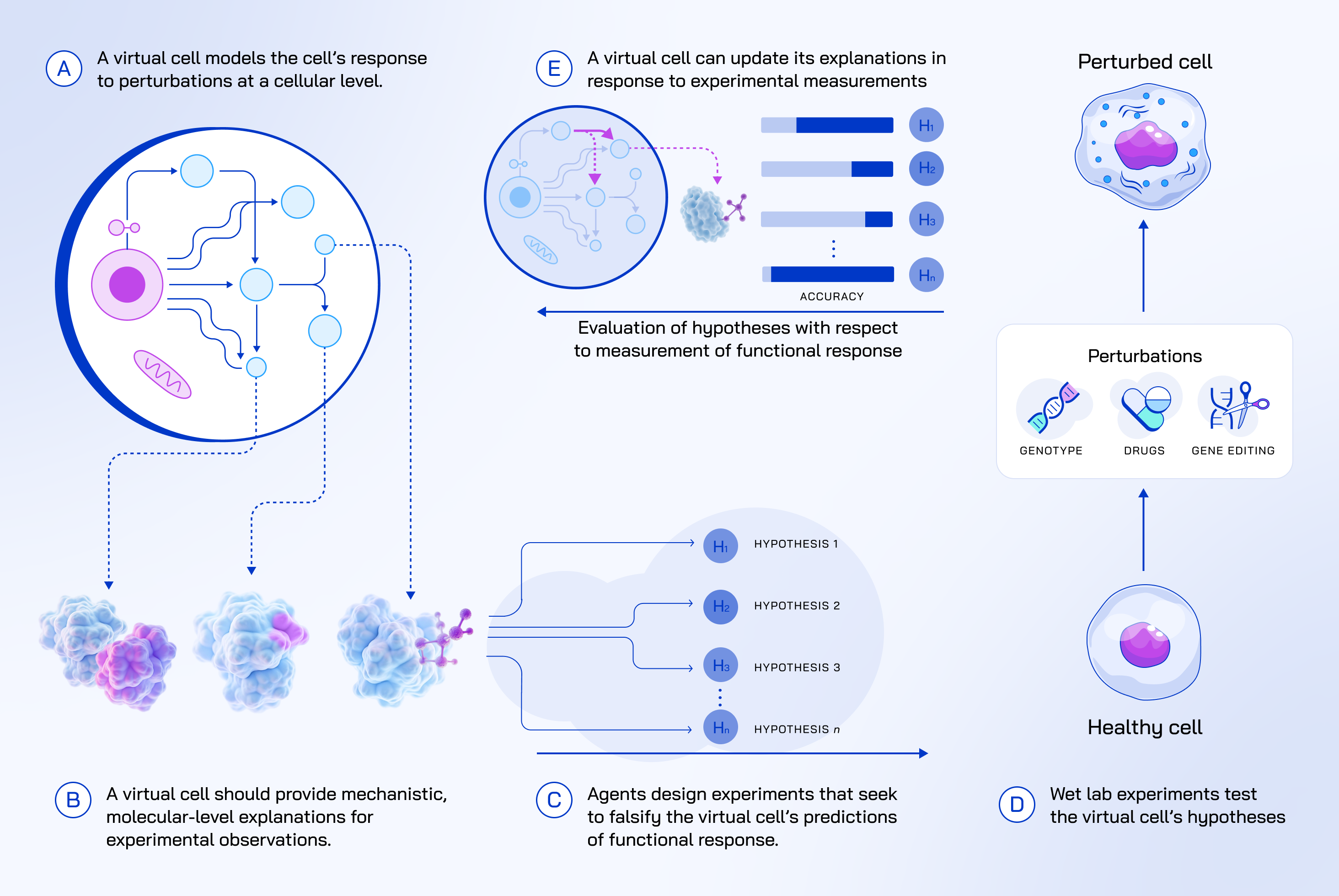}
    \caption{\small\textbf{A vision for virtual cells: the Predict-Explain-Discover capabilities in action.} \textbf{A.} Virtual cells \textit{predict} the functional response of a cell to a perturbation. \textbf{B.} Virtual cells explain the prediction via mechanistic descriptions of key molecular interactions. \textbf{C.} Agents design experiments that seek to falsify virtual cell predictions in order to \textit{discover} novel biology. \textbf{D.} These experiments are executed in real-world labs to test the generated hypotheses. \textbf{E.} Virtual cells are updated whenever a hypothesis is falsified, closing the loop between prediction and measurement. Together, these capabilities enable therapeutic discovery.}
    \label{fig:vc-vision}
\end{figure}

We too share the objective of predicting holistic functional response, but our vision extends further: we seek to build virtual cells that can also infer molecular-scale interactions from cellular-scale observations.  Thus, a virtual cell should \textit{explain} cellular-level observations in terms of molecular-level mechanisms, without resorting to whole-cell simulation---the second of the P-E-D capabilities. Furthermore, virtual cells must also be capable of generating testable hypotheses to \textit{discover} novel treatments for patients---the third of the P-E-D capabilities. See \Cref{fig:vc-vision} for an overview of how the P-E-D capabilities of virtual cells work together to drive therapeutic discoveries. In the sections that follow, we expand on each core capability and introduce a corresponding \textit{design principle} to guide virtual cell development.

\subsection{Virtual cells should predict functional responses}\label{sec:vision-predict}

A cellular functional response refers to the change in behavior, state, or activity of a cell in response to stimuli, such as compounds, genetic modifications, or environmental changes. These responses can manifest across many biological levels, including gene expression, protein activity, signaling pathways, morphology, proliferation, and secretion. In recent years, our ability to generate data capturing aspects of these functional responses has improved dramatically, due in large part to advances in high-throughput omic techniques---genomic, transcriptomic, proteomic, phenomic, metabolomic, epigenomic, etc.---at increasingly fine resolution, including at the single-cell and even single-molecule levels. These datasets are often \textit{interventional}, meaning they capture a readout from a cell after intervening on it, usually with one or more \textit{perturbations}---molecular tools designed to alter cellular function either temporarily or permanently. Examples of perturbations include gene knockout \citep{perturb-seq}, gene overexpression \citep{joung2017genome}, treatment with small molecules \citep{ye2018drug}, extracellular stimulation with soluble factors \citep{cuccarese2020functional}, and infection with viral agents \citep{heiser2020identification}. 

Taken together, these interventional datasets provide an extensive, multimodal view of how cells respond to perturbations, and learning to predict these functional outcomes, by training AI/ML models on these data directly rather than attempting whole-cell simulation, offers a practical path for building therapeutically-relevant virtual cells. Such predictive models\footnote{To be useful, such virtual cells also need to generalize well across perturbation types, cell types, and assay conditions, while maintaining robustness to batch effects and other technical noise.} would be surrogates for experimental assays and enable the testing of therapeutic hypotheses \textit{in silico}, introducing a new paradigm of rational drug design at the whole-cell level rather than at the level of individual biomolecules \citep{hopkins2008network,moffat2017opportunities,rafelski2024establishing}. Therefore, the first key capability we envision for virtual cells is to accurately \textit{\textbf{predict the functional response of cells to perturbation}}.

\begin{figure}[tb]
\centering
\includegraphics[width=0.9\textwidth, valign=t]{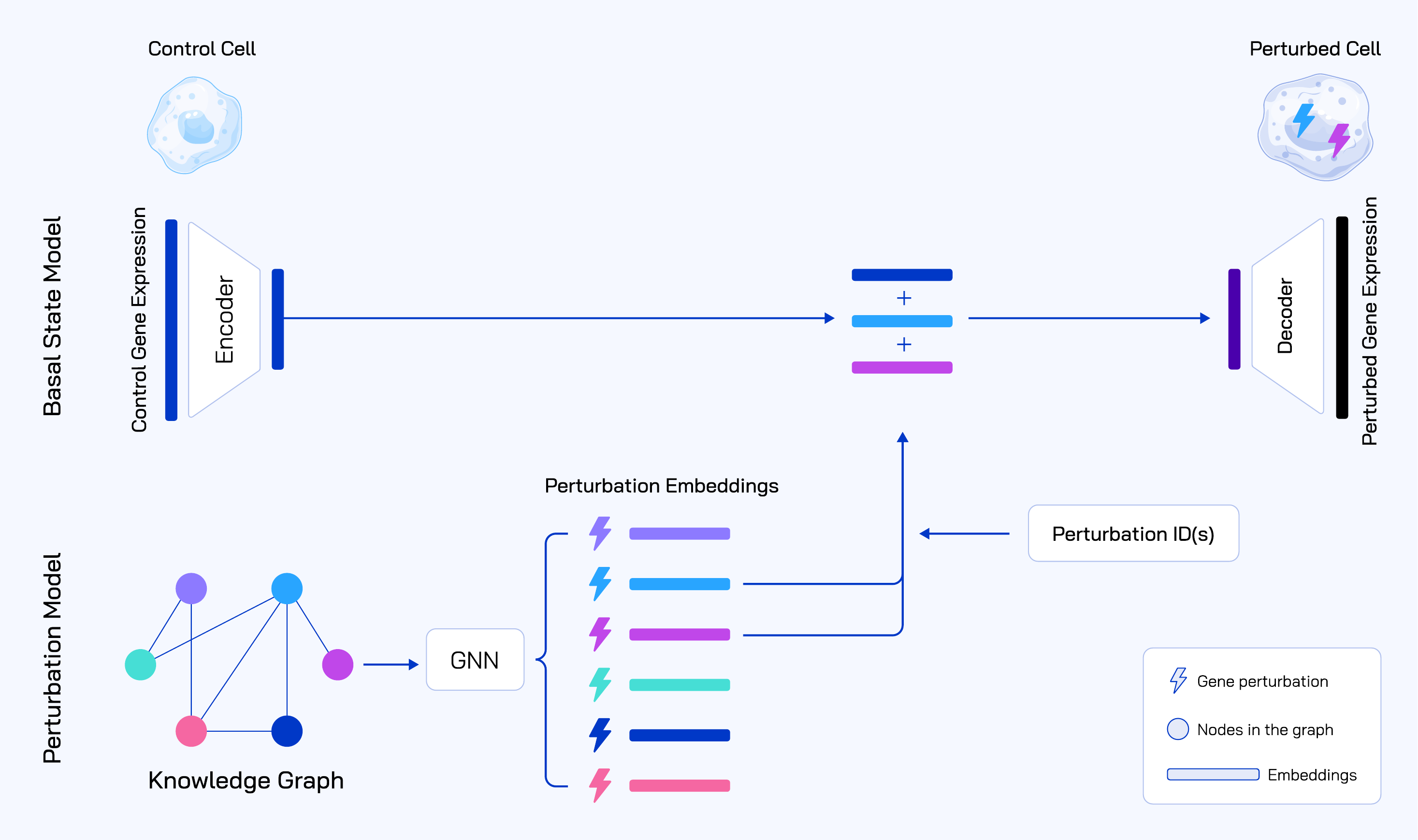}
\caption{\textbf{Virtual cells should predict relative changes.} Models like TxPert~\citep{wenkel2025txpert} predict the change in the readout (e.g. gene expression) of a perturbed cell given the readout of a control cell. Doing so allows the model to focus on learning the effect of the perturbation on a given control state, which provides the context for the prediction.}
\label{fig:vc-relative-change}
\end{figure}

\paragraph{Design principle 1: Predict relative changes} To predict functional response, virtual cell predictions should be expressed as changes relative to the state of the cell to which the perturbation was applied; i.e., the prediction is conditioned on the state of the cell prior to perturbation (see \Cref{fig:vc-relative-change}). 

Conditioning on the initial state of the cell ensures that models focus on features directly relevant to the perturbation and its context. 
Ideally, virtual cells are trained on temporal trajectories from the same biological sample, measured before and after a perturbation, thereby enabling them to learn how the perturbation alters the cell's developmental path. Thus the model would learn that an effect is only expected in certain cellular contexts where the perturbation target is active, and that its magnitude depends on the state of the cell at the time of intervention. 

While such trajectories are rare\footnote{Many common cellular assays destroy samples in order to obtain readouts (e.g., RNA-seq methods require lysing cells in order to capture RNA). Live cell assays like brightfield microscopy can preserve samples across timepoints.}, most interventional datasets do include \textit{negative control} samples\footnote{In practice, negative control samples are not simply untreated samples. Instead, they are treated with a perturbation of the same type which is expected to induce only the shared effects of that perturbation type, so that the only difference between negative controls and perturbed samples is the targeted effect of the perturbation. For example, CRISPRn controls may use guides targeting introns, which does not (intentionally) knock out any gene, but does induce the common effects of DNA cutting (e.g., DNA damage response).} that approximate the unperturbed distribution. Using these controls further helps isolate the targeted effect from conserved cellular programs, such as housekeeping gene expression and homeostatic machinery.

This formulation not only improves robustness but also naturally fits the interventional nature of most available datasets. Although experimental metadata are often sparse, and small contextual differences can have significant effects on outcome\footnote{Such differences in experimental conditions are often the origin of batch effects.}, conditioning predictions on the observed initial state allows virtual cells to absorb such variability. This reduces the need for aggressive pre-processing or batch correction and enables training across heterogeneous datasets with minimal harmonization. %

\subsection{Virtual cells should explain functional responses in terms of molecular mechanisms}\label{sec:vision-explain}

While accurate prediction is essential, virtual cells must also provide \textit{explanations}: structured, testable accounts of how perturbations give rise to observed cellular outcomes. Such explanations are critical for guiding experimental design, generating hypotheses, supporting therapeutic decisions, evaluating model trustworthiness, reasoning about biological system behavior, and identifying actionable points of intervention.

In theory, quantum mechanical (QM) approaches such as density functional theory (DFT) can accurately model biomolecular systems all the way down to the electronic level. However, they are computationally prohibitive for large systems and often fail to capture critical interactions like van der Waals forces~\citep{grimme2016dispersion,caldeweyher2019generally}. Even simulating a few nanoseconds of molecular dynamics for modest systems can consume entire compute-days~\citep{Cole_2016}. Force-field-based methods improve efficiency at the cost of physical realism, limiting their reliability for modeling complex or long-timescale biochemical processes. Sparse experimental data further compound these limitations. Structural biology techniques like X-ray crystallography and cryo-EM and their deep learning counterparts such as AlphaFold and related approaches~\citep{jumper2021highly,abramson2024accurate,wohlwend2024boltz,chai2024chai} provide high-resolution but static snapshots of biomolecular structures. Although time-resolved methods like FRET~\citep{sekar2003fluorescence} and trEM~\citep{amann2023frozen} offer dynamic insights, they remain too costly and low-throughput to scale. Altogether, these computational, methodological, and empirical constraints make simulating an entire eukaryotic cell from first principles a distant goal.

Nonetheless, targeted atomistic simulations remain highly effective for modeling tractable events such as ligand–receptor binding, protein–protein interactions, allosteric regulation, and local conformational shifts. These applications are already central to drug discovery and provide essential mechanistic anchors for interpreting cellular responses~\citep{anderson2003process,sliwoski2014computational,de2016role,durrant2011molecular}.

\begin{figure}[tb]
\centering
\includegraphics[width=0.85\textwidth, valign=t]{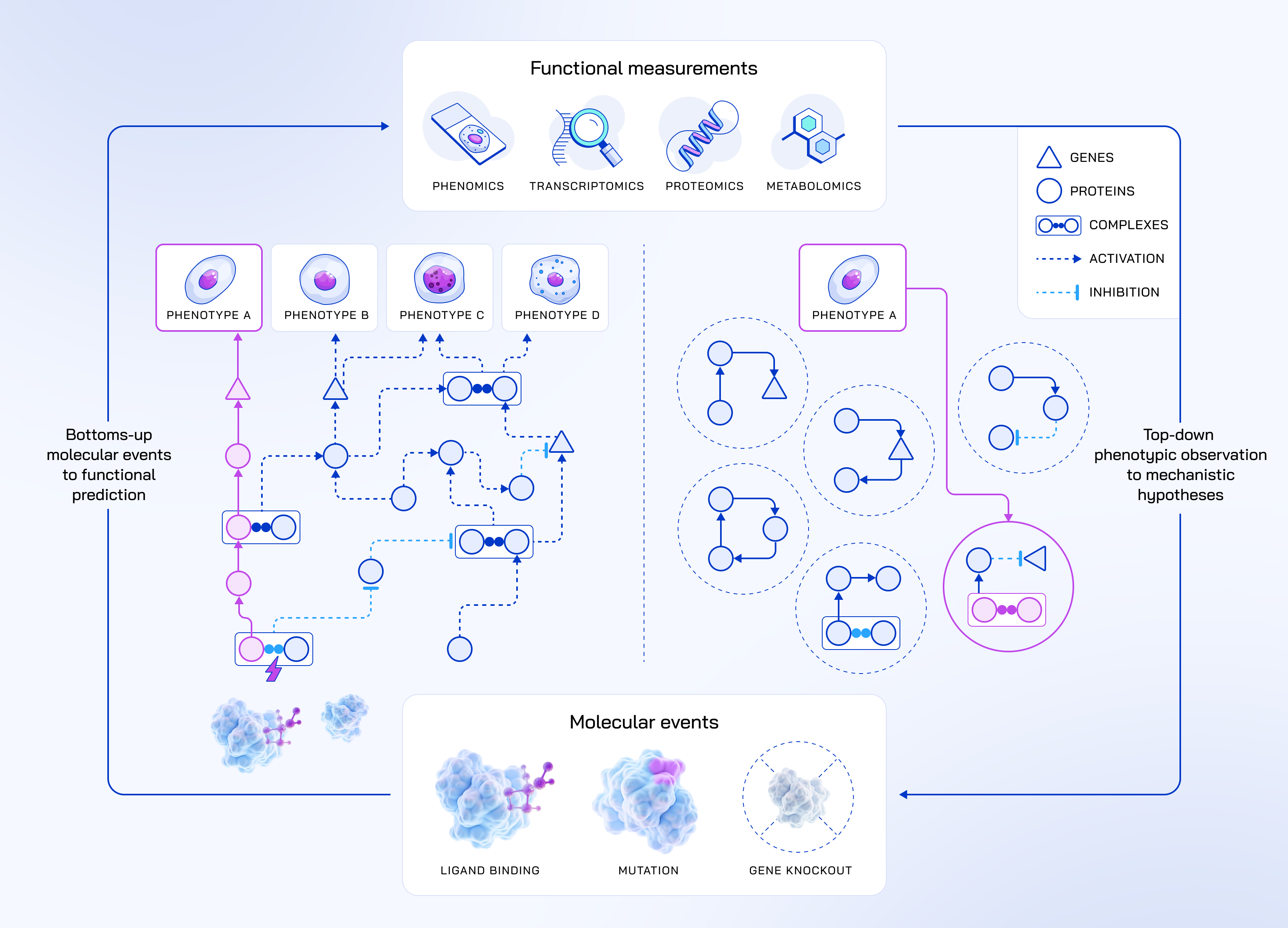}
\caption{\textbf{Virtual cells bridge top-down functional prediction with bottom-up molecular explanations.} A virtual cell's functional predictions should be informed by omic-level observations and constrained by physical molecular interactions. Together these combine to provide a mechanistic model of a cell, propagating the effect of a perturbation from a change in molecular interactions through the affected pathways to the observed functional response.}
\label{fig:vc-atom-to-cell}
\end{figure}

Here too, AI/ML offer a path to scale these capabilities. ML-based interatomic potentials (MLIPs) trained on QM data can achieve near-quantum accuracy while accelerating \textit{ab initio} simulations by several orders of magnitude~\citep{martin2024overview,mann2025egret1pretrainedneuralnetwork}. Generative models~\citep{jing2024generative,pang2025deeppath} can simulate conformational transitions, upsample sparse MD trajectories, and design molecules under structural constraints. Incorporating these tools into virtual cells enables time-resolved, mechanistically grounded modeling of key molecular events, bridging top-down functional predictions with bottom-up molecular causes (see \Cref{fig:vc-atom-to-cell}).

Such integration unlocks new capabilities for perturbation modeling, causal inference, and the interpretation of dynamic cell state transitions. ML-accelerated atomistic simulation thus plays a key explanatory role: anchoring predictions in molecular reality and supporting scalable, testable hypothesis generation.

Therefore, the second key capability we envision for virtual cells is to \textit{\textbf{explain the functional response of cells to perturbation}} in terms of key molecular mechanisms.

\paragraph{Design principle 2: Explain perturbations as dynamic changes to key biomolecular interactions}  
To explain functional responses, virtual cells must model how perturbations dynamically alter the structure, activity, or strength of biomolecular interactions. We adopt the systems biology view that cells function through organized, dynamic networks of interacting molecules, structured into pathways that coordinate information flow and control. Perturbations shift these networks: proteins bind differently, conformations change, and regulatory circuits rewire.

Virtual cells should frame these effects as cascades of changes to a minimal, coherent set of \textit{key interactions}---binding affinities, post-translational modifications, transcriptional regulation, signal propagation---sufficient to explain observed outcomes. 
These hypotheses can be informed by ML-based structural biology tools such as AlphaFold~\citep{abramson2024accurate} and Boltz~\citep{wohlwend2024boltz}, and refined through targeted atomistic simulation.

This mechanistic framing is especially valuable for therapeutic discovery. Virtual cells grounded in structural modeling can identify cryptic pockets, allosteric sites, or conformational vulnerabilities. When paired with generative molecular design tools~\citep{winnifrith2023generative,du2024machine,noutahi2024gotta,roy2023goal,cretu2025synflownet}, they can propose plausible, mechanism-based interventions, thus accelerating the design–simulate–test cycle.

Identifying which interactions are \textit{key} remains a modeling challenge. A pragmatic strategy is to prioritize candidate mechanisms that improve predictive performance or yield falsifiable hypotheses. For instance, a virtual cell might suggest that inhibiting a kinase alters transcription via a defined signaling cascade, testable through CRISPR knockouts, pathway reporters, or perturbation screens.

Importantly, we do not require virtual cells to replicate canonical biological pathways. Instead, they may uncover alternative structures that better fit the data, improve generalization, or offer more interpretable insights. Explanations should therefore be evaluated not by their agreement with textbook biology, but by their ability to support causal reasoning, guide experiment design, and generate testable hypotheses.

\subsection{Virtual cells should discover therapeutically-actionable biology}\label{sec:vision-discover}

Virtual cells are an essential component of a broader vision for improving drug discovery. They serve as world models of the cellular biology we observe in the ``real world" via lab experiments. As such, they embody our best understanding of human cellular physiology and pathology, encapsulating the mechanistic effects of the network of biomolecular interactions that underlies cellular function. 

A well-designed virtual cell should thus not only recapitulate known biology but also act as a hypothesis engine for discovering new biology. By generating responses across diverse perturbations and cellular contexts, virtual cells can propose novel mechanisms, identify promising interventions, uncover therapeutic opportunities, and prioritize experiments that yield the greatest insight.

This motivates a lab-in-the-loop paradigm, where virtual cells iteratively suggest experiments that are maximally informative. These experiments test predictions, falsify incorrect hypotheses, and update the model. Conceptually, a virtual cell becomes a testable theory of human cell physiology, continuously refined through empirical feedback~\citep{popper2005logic,Corfield_2009}. Therefore, the third key capability we envision for virtual cells is to \textit{\textbf{discover novel biology}} through iterative, hypothesis-driven experimentation.

\paragraph{Design principle 3: Lab-in-the-loop falsification to align virtual cells with the real world}\label{sec:vision-discover-falsification}

We envision virtual cells as being initially trained on available data, grounding them in experimentally-observed cellular behavior. Once trained, however, they could be paired with active learning and sequential model-based optimization techniques \citep{jain2023gflownets,pauwels2014bayesian,sverchkov2017review} to design and prioritize experiments that are most informative for improving performance, expanding functional coverage to new cellular contexts and increasingly complex perturbation combinations. 

This sets the stage for a continual lab-in-the-loop improvement of virtual cells. Conceptually, a virtual cell can be considered a theory of human cellular physiology and pathology, one that generates a diverse set of statements about the real world that we observe via experiments. Following an active learning interpretation of Popper’s theory of scientific discovery \citep{popper2005logic}, we can imagine one or more agents, whether human or artificial, choosing hypotheses for testing, and updating the virtual cell ``theory'' whenever a claim is falsified \citep{Corfield_2009}. Actively seeking experiments that falsify the current virtual cell could lead to surprising and interesting novel biology \citep{zuheng2024automateddiscoverypairwiseinteractions}.

\begin{figure}[tb]
\centering
\includegraphics[width=\textwidth]{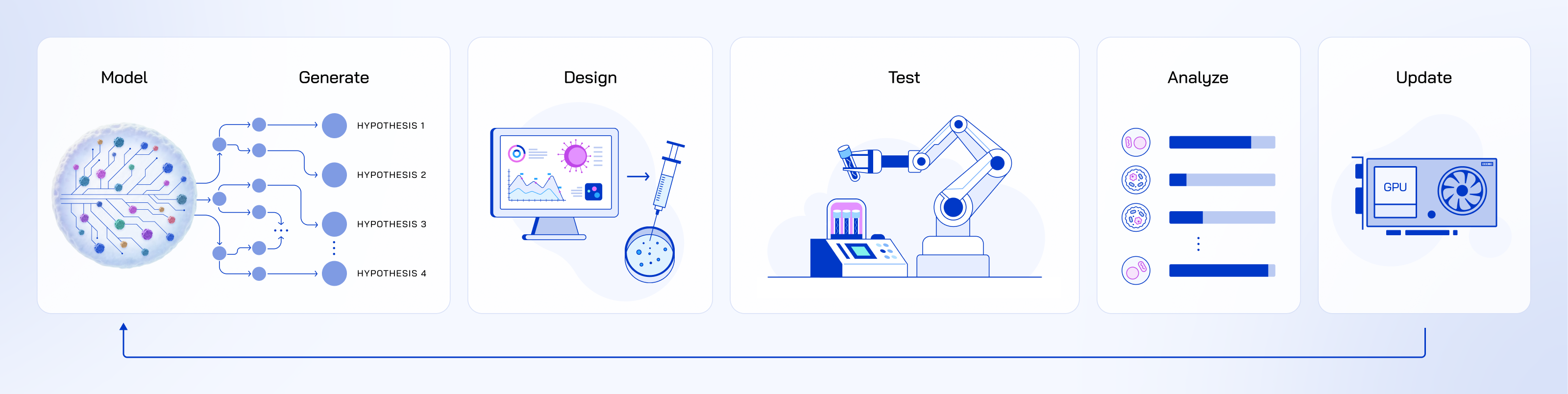}
\caption{\textbf{Virtual cells are falsifiable theories of cellular biology.} Virtual cells are world models of cellular biology, initially trained on existing experimental data, then guiding the selection of new experiments that are most informative for improving their performance. This iterative, lab-in-the-loop approach can be turned into a program of ``theory falsification'' to generate novel drug discoveries.}
\label{fig:vc-world-model}
\end{figure}

Furthermore, by conditioning the generation of hypotheses on a particular disease, the agent could steer this active improvement toward discovery of potential treatments more efficiently \citep{neporozhnii2025efficient}. From a drug discovery perspective, such virtual cells could lead to a radical change in how programs are run, shifting from cycles of \textit{design--make--test--measure} to \textit{design--simulate}. \Cref{fig:vc-world-model} illustrates these concepts.

\subsubsection{A path to scientist AIs}\label{sec:vision-discover-scientist-ais}

If we imagine the agents operating within this lab-in-the-loop framework as autonomous agentic systems capable not only of the conditional generation of hypotheses using virtual cells, but also:
\begin{itemize}
\item prioritizing hypotheses for falsification,
\item designing experiments to efficiently test those hypotheses,
\item orchestrating the execution of experiments,
\item analyzing the experimental outcomes, and
\item integrating these into the virtual cell for iterative refinement,
\end{itemize}
then the falsification framework described in the previous section becomes a plausible path toward developing genuine \textit{scientist AIs} capable of making novel discoveries and fundamentally transforming drug discovery workflows. Indeed, we envision a future in which such scientist AIs, with access to high-fidelity virtual models and the ability to be prompted in natural language, could accept a query such as  ``propose a set of mechanistically distinct compounds predicted to be effective in treating patients with the following clinical and laboratory parameters (see attached table), which may be representative of a disease that is part of a heterogeneous group of diseases often referred to as stage II non-Hodgkin's lymphoma''. From there, they could iteratively and autonomously work toward the discovery objective, ultimately generating drug candidates with high likelihood of translating successfully to human patients. 

Realizing this vision will require systematic, biologically grounded benchmarks that can track progress across the core capabilities of virtual cells.

\section{Toward Rigorous and Biologically-Grounded Benchmarks for Virtual Cells}\label{sec:benchmarking} %

In designing virtual cells, it is essential to consider their downstream applications and ensure they can be systematically evaluated. Benchmarking is not an afterthought, but a fundamental constraint that must shape how virtual cells are built, trained, and validated. Driving measurable progress across different approaches will require a suite of benchmarks covering a wide range of functional responses, cellular contexts, and perturbations, with objectives spanning prediction, explanation, and discovery. To achieve this, benchmarks must capture both predictive accuracy and biological relevance, while aligning evaluation with the broader objectives of understanding and modulating cellular states for therapeutic discovery. Four major aspects must be systematically covered:

\paragraph{Functional responses.} We expect functional responses to typically be measured as observed changes in omics-level readouts, like changes in gene expression captured by scRNA-seq transcriptomic assay or changes in morphological features as measured in a brightfield phenomic assay, following one or more perturbations. While neither gene expression changes nor morphological shifts alone fully capture a cell’s functional response, they provide relevant, measurable views of cellular behavior. To move closer to capturing true response, benchmarks should aim to integrate information across multiple modalities, rather than relying solely on a single type of readout that may reflect modality-specific biases or noise. We must leverage the available data and construct meaningful benchmarks from these partial but complementary measurements \citep{wognum2024call,tossou2024real}.

\paragraph{Cellular contexts.} Cellular contexts can be characterized by a variety of biological states: different cell types, stages of division, differentiation, metabolism, signaling activities, spatial organization, and other physiological descriptors. In addition, the surrounding microenvironment, such as the tissue architecture, neighboring cell types, and extracellular signaling milieu, plays a critical role in shaping cellular responses. Because cellular responses to perturbations are shaped by both intrinsic state and external conditions at the time of intervention, it is essential that we evaluate our virtual cells across a wide range of physiologically-relevant settings.  To support our stated therapeutic and translational goals, benchmarks should test generalization across a wide array of disease-relevant contexts.

\paragraph{Perturbations.} The term \textit{perturbation} refers to a wide array of tools designed to engage or disrupt mechanisms in and on the cell, altering its function either temporarily or permanently. CRISPR technologies knock out genes, silence translation, or increase transcription. Chemical compounds bind proteins and alter their conformations affecting downstream activities. Soluble factors such as cytokines initiate extracellular signaling cascades, while antibodies can trigger immune responses. Collecting readouts across a wide range of perturbations gives us a broad view of the inner workings of cells. Importantly, some perturbation mechanisms, particularly in specific cellular contexts, have been well characterized through decades of research. This prior knowledge can be leveraged when building benchmarks, enabling the integration of perturbations with varying levels of mechanistic certainty. Virtual cell models that accurately predict functional response across this spectrum will provide invaluable insights into how to modulate disease-associated cellular states, resulting in testable therapeutic hypotheses.

\paragraph{Predict-Explain-Discover.} As described in our vision for virtual cells (Section \ref{sec:vision}), we want to build virtual cells that not only accurately predict functional responses, but explain those responses mechanistically and drive novel drug discovery. Thus our benchmarks must assess all three of these capabilities. Predictions are relatively easy to benchmark because any new experiment generates data that could be used to benchmark a virtual cell. In contrast, evaluating explanations and assessing novelty are more difficult, as it requires assessing statements about unobserved biological mechanisms that would typically demand multiple specialized assays. While we can benchmark explanations and discovery with known biology, such efforts need to be very carefully controlled to avoid data leakage, particularly in light of the growing reliance on literature-derived information as a way of constructing gene embeddings \citep{chen2024genept}.

Following these principles, we describe a framework of capabilities we expect virtual cells to demonstrate as the field advances, and recommend that all benchmarks designed for virtual cells are intentionally associated with one or more of these capabilities. Doing so will help to better mark progress, differentiate between virtual cell models, and guide ongoing research efforts.

\Cref{tab:vc-capabilities} summarizes this framework of capabilities and gives examples of how they could impact drug discovery and our understanding of biology. While these capabilities are organized along increasing levels of complexity, we acknowledge that they do not need to be achieved in the order presented here, nor do we claim that this list is exhaustive. We note that our framework primarily captures the Predict and Explain capabilities of virtual cells, which we believe are the core enablers of discovery. As described in earlier sections, we view Discover not as a separate axis to benchmark in isolation, but as the natural consequence of predictive and explanatory models applied to therapeutic contexts. In particular, discovery is best evaluated through a model’s ability to generate testable and actionable hypotheses, an aspect that depends on but extends beyond the capabilities presented here.

\begin{table}[tb]

\centering

\caption{\textbf{A framework of virtual cell capabilities.} Each capability unlocks specific aspects of biological understanding and enables downstream applications.}
\label{tab:vc-capabilities}

\footnotesize

\renewcommand{\arraystretch}{1.2}

\begin{tabular}{>{\raggedright\arraybackslash}p{3.1cm} >{\raggedright\arraybackslash}p{5.6cm} >{\raggedright\arraybackslash}p{5.6cm}}

\toprule

\textbf{Predicts response\ldots} & \textbf{Understandings unlocked} & \textbf{Applications} \\

\midrule

\multicolumn{3}{c}{\textit{Observational capabilities}} \\
\addlinespace[0.5em]

For specific biology
& Biomolecular features
& Predicting protein localization; reconstructing higher-resolution readouts from bulk or lower-resolution assays \\

\addlinespace[0.5em]

For core biology 
& Modality-invariant biological processes (e.g., transcription–translation coupling)
& Predicting pathway-level effects of perturbations;  replacing expensive modalities with cheaper ones (e.g., transcriptomics with imaging) \\

\addlinespace[0.5em]

For unobserved biology
& Latent molecular states (e.g post-translational modifications, hidden regulators)
& Inferring unseen regulatory nodes; grouping genes and compounds into higher-order functional modules (e.g., pathways) \\

\addlinespace[0.5em]
\hline
\addlinespace[0.5em]
\multicolumn{3}{c}{\textit{Contextual capabilities}} \\
\addlinespace[0.5em]

Given intrinsic context 
& Effect of genotype, cell type, biomolecular state. 
& Generalizing across cell types, genotypes, and developmental states \\

\addlinespace[0.5em]

Given extrinsic context 
& Influence of assay design, local microenvironment, and intercellular signaling 

& ADME-T prediction; modeling effects of experimental conditions and local tissue context (e.g., proximity-driven signaling, tissue-level response) \\

\addlinespace[0.5em]
\hline
\addlinespace[0.5em]
\multicolumn{3}{c}{\textit{Explanatory capabilities}} \\
\addlinespace[0.5em]

Over time 
& Temporal integration of molecular and cellular dynamics
& Treatment response over time \\

\addlinespace[0.5em]

Causally 
& Mechanistic understanding of regulation and interaction
& Mutation effect prediction; causal modeling \\

\bottomrule

\end{tabular}
\end{table}

\subsection{Observational capabilities}\label{sec:benchmarking-capability-observational}

As performance moves beyond simple baselines, we want to assess how much a virtual cell's predictions can leverage the different types of readouts that are provided. While we advocate for modality-agnostic evaluations where possible (see Appendix \ref{sec:benchmarking-considerations} for an extended discussion on benchmarking considerations), in practice this means measuring response predictions with respect to available modalities. These capabilities, therefore, begin with the specific biology accessible within a single modality, then expand to focus on predicting functional responses for core biology, which should be observable across all modalities, and unobserved biology, which may not be directly observed but is nonetheless known to play a role.

\paragraph{Predicts response for specific biology.} The scope of this capability is relatively narrow, measuring the extent to which a virtual cell predicts functional response from single modalities. Nevertheless, since ground-truth examples used for evaluation will often come from multiple experiments, models that perform well here will likely need to predict across differences in context (see \Cref{sec:benchmarking-capability-contextual} for more on contexts) in order to deal with \textit{batch effects} and other sources of variation across these datasets.

\paragraph{Predicts response for core biology.} Here we broaden the scope to functional responses related to the central dogma. There are a set of core biological processes that should appear regardless of the modality in which they are observed. For example, for genes that maintain basic cellular functions---actin, GAPDH, ubiquitin, etc.---the presence of their transcripts implies the presence of their protein products, and they should be observable in both transcriptomics and proteomics. Thus this capability assesses the extent to which a virtual cell can capture fundamental biology.

\paragraph{Predicts response for unobserved biology.} This capability broadens the scope even further to predicting functional responses that are not directly observed in any of our modalities, but are nonetheless expected based on our knowledge of biology. For example, predicting that an RNA but not the corresponding protein is produced (long non-coding RNAs), or that an RNA rather than a protein catalyzes a biological reaction (ribozymes) is biologically plausible. Similarly, a given readout may not explicitly capture phenomena such as post-translational modifications but should be accounted for nonetheless. The extent to which we can accurately predict into these “dark” areas of biology will help us understand which biological processes are represented in each modality, which subsets are most informative for making a particular prediction, and provide evidence that the model is capable of producing novel insights.

\subsection{Contextual capabilities}\label{sec:benchmarking-capability-contextual}

Cells do not exist in a vacuum but within various levels of context determined by details of the cell itself, environmental factors, and the milieu of surrounding cells and biological constituents. These capabilities, therefore, measure the ability to predict functional response across a broadening set of biological contexts. Note that we could continue to describe capabilities at higher levels of biological organization, but those would be more appropriate for benchmarking correspondingly higher-levels of virtual models, and thus omit them here.

\paragraph{Predicts response given intrinsic context.} By \textit{intrinsic context}, we mean all those factors describing the biology of the cell: genotype, cell type, cell cycle phase, pathway activation, levels of key biomolecules including RNA, protein, and ATP. Virtual cells that account for these biological factors will accurately predict response across a wide range of cellular contexts. For example, cell types differ significantly in gene expression patterns, morphology, and regulatory programs. Virtual cells must account for these differences to accurately predict responses adapted to each cellular identity.

\paragraph{Predicts response given extrinsic context.}  By \textit{extrinsic context}, we refer to all external factors influencing cellular behavior, including assay design (e.g., growth media, temperature, incubation time, reagent concentrations) as well as the spatial and cellular microenvironment. This includes interactions with nearby cells, such as contact inhibition, gap junction signaling, or paracrine communication, which can modulate cell state and response.  Virtual cells that account for these confounding factors will generalize across different experimental conditions by modeling the complex interactions between cellular response and environment. We note that the effects of extrinsic context on a cell are often mediated through the biological factors of its intrinsic context, suggesting that this capability encompasses the previous one.

\subsection{Explanatory capabilities}\label{sec:benchmarking-capability-explanatory}

As outlined in the vision (Sections \ref{sec:vision-predict}, \ref{sec:vision-explain}, and \ref{sec:vision-discover}), %
the most useful versions of virtual cells offer a mechanistic understanding of cellular biology in order to provide the evidence needed to understand therapeutic hypotheses. These capabilities, therefore, benchmark progress toward mechanistic explanations of a virtual cell's predictions.

\paragraph{Predicts response over time.} Cells are dynamical systems in a continual state of flux, converting chemical energy into work to maintain their internal homeostasis while responding to external stimuli. Although an observed response in any modality implicitly captures the notion of time, this virtual cell capability requires the time component to become explicit in predicting response. Accurately predicting the time-course of response will likely require a virtual cell to model how the interaction of key biomolecules over short timescales integrates over time to produce the response at longer timescales, potentially providing mechanistic insights that can be leveraged in drug discovery and other applications.

\paragraph{Predicts response causally.} Knowing the reasons for \textit{why} a particular response occurs is much more useful than simply predicting that it does. Identifying causal relationships between observed variables and building from them a mechanistic model allows virtual cells to more accurately generalize their predictions out-of-distribution, and move toward understanding the plethora of counterintuitive nonlinear phenomena we observe in biology, e.g., epistatic interactions like synthetic lethality. Causal learning, however, is a notoriously difficult problem in general, and particularly so in the high-dimensional omics datasets available within biology.

\subsection{Performance levels for virtual cells}\label{sec:benchmarking-levels}

To further drive progress, we can organize the capabilities described above into performance levels that represent key milestones toward building increasingly powerful virtual cells.

We propose the following three levels as an initial framework for evaluating progress:

\textbf{VC level 1.} Predicts and explains functional response for specific biology, given intrinsic and extrinsic context, at a single timepoint.

\textbf{VC level 2.} Predicts and explains functional response for core biology, across environmental variation, and over time.

\textbf{VC level 3.} Predicts and explains functional response for unobserved biology, in spatially organized systems, and with causal interpretability.

While the exact definition and composition of these levels may need to be adapted over time, we hope that adopting performance milestones for virtual cells will help organize the research community around common standards and research directions, and ultimately lead to the realization of powerful virtual cells sooner than otherwise.

\section{A framework for building virtual models at higher levels of organization}\label{sec:higher-organization}

In this perspective, we have outlined a vision for therapeutically-relevant virtual cells that predict functional responses with respect to multiple modalities, explain these in terms of molecular mechanisms, and generate novel discoveries through lab-in-the-loop experimentation. We have taken the view that such virtual cell models should combine omics data and atomistic simulation to \textit{predict}, \textit{explain}, and \textit{discover}---that is, to model the functional response of cells to perturbation, reveal the underlying molecular mechanisms, and generate novel and falsifiable biological insights through iterative hypothesis testing and refinement. To support this development, we have also outlined key design principles for virtual cells and proposed a framework for the adoption of biologically-meaningful benchmarking standards to help shape the research community's efforts.

\begin{figure}[htb]
    \centering
    \includegraphics[width=0.85\textwidth, valign=t]{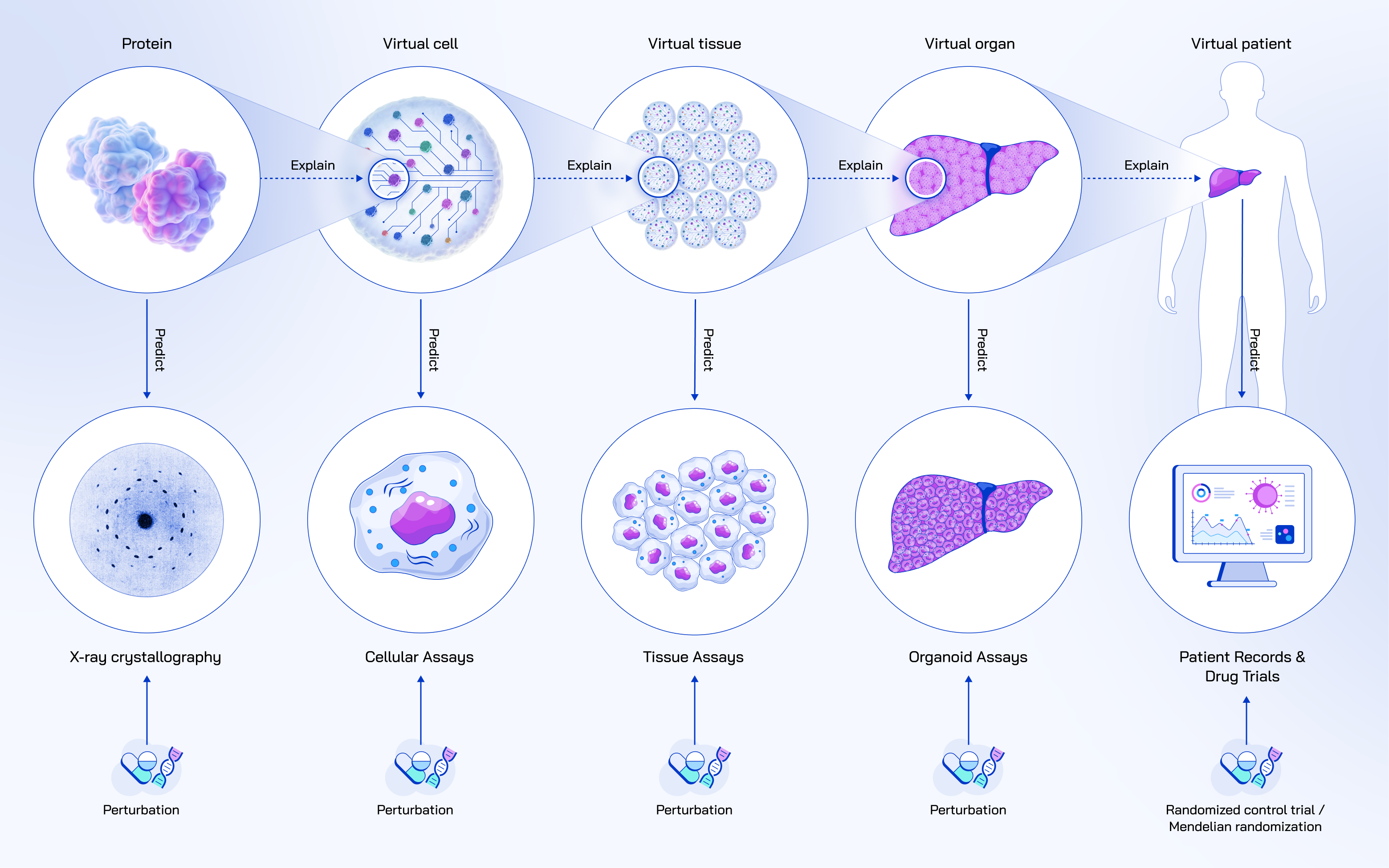}
    \caption{\small\textbf{From virtual cells to virtual patients.} By focusing on predicting functional response and explaining in terms of lower-level mechanisms, we can continue to build virtual models that represent higher levels of biological organization, eventually arriving at a virtual patient.}
    \label{fig:virtual-patient}
\end{figure}

Our primary objective in developing virtual cells as described herein is to accelerate the drug discovery process, reduce development costs, and enable more precise therapeutic intervention. While virtual cells alone will not solve every challenge, we see them as prototypical examples for virtual models at higher levels of biological organization, from virtual tissues to virtual organs and ultimately virtual patients. Importantly, we believe that virtual models at every level of organization can be built in a manner similar to what we described for virtual cells---by training models with Predict-Explain-Discover capabilities on data that suitably describes functional response at the respective level (see \Cref{fig:virtual-patient} for an illustration of these ideas). 

Each step toward higher-order modeling extends the Predict-Explain-Discover capabilities of virtual models into more structured and interconnected systems. While each level could build on the one below---tissue-level predictions emerge from virtual cells interacting in space and time; organ-level behavior depends on the coordination of multiple tissue types; patient-level outcomes reflect the integrated behavior of organ systems shaped by genetics, disease, and treatment history---this progression does not need to be strictly sequential. Indeed, advances in biotechnology such as organs-on-chips \citep{danku2022organ} and ``\textit{in vivo} omics''~\citep{vium_lim2017development,vium_baran2021}  can provide unprecedented access to structured biological responses at the tissue, organ and whole-organism levels. Combined with the increasing availability of multimodal and anonymized patient-specific profiling, these developments make it increasingly feasible to model higher-order biological organization directly. 

At the tissue level, groups of interacting cells model a collective response to perturbations, predicting spatially-distributed effects such as signal propagation or local response heterogeneity, explain how intercellular communication shapes tissue phenotype, and could generate hypotheses about how spatial context modulates drug sensitivity \citep{yuan2016spatial}.

Virtual organs add another layer of complexity: they must model structured spatial organization and long-range signaling (e.g., hormonal or vascular). Predictions at this level recapitulate outputs from multiple tissue types, explanations involve anatomical dependencies and interactions between regulatory axes, while discovery may focus on falsifying hypotheses around organ-specific responses, immunogenicity, toxicity patterns, or compensatory dynamics that mask intervention effects \citep{sontheimer2019modelling,marx2020biology}.

At the patient level, models integrate across organs to simulate whole-body responses to interventions. These include predicting clinical outcomes such as biomarker trajectories, disease progression, or treatment efficacy over time; and explaining sources of inter-patient variability arising from differences in genetics, physiology, and environment. Virtual patient models could generate testable hypotheses for personalized treatment strategies, for example, identifying subpopulations more likely to benefit from a given therapy, or proposing individualized dosing regimens based on predicted pharmacokinetic and pharmacodynamic profiles~\citep{minichmayr2024model}.

While only a vision today, it is not unreasonable to expect that we can make progress towards these objectives based on the key advances we highlighted in \Cref{sec:intro-advances}. Successfully delivering this progress would enable personalized medicine at unprecedented levels of accuracy and scale, allowing for the development of tailored therapeutic interventions based on patient-specific virtual simulations.

In the meantime, we invite the research community to engage with this perspective by refining, challenging, and extending the ideas presented herein, as we collectively progress toward a new way of doing therapeutic discovery. We hope to see the adoption of rigorous benchmarking standards as the research community working on virtual cells grows, and hope that the principles and capabilities described herein can be a useful tool in steering our collective efforts.

\section{Acknowledgments}
The perspective presented in this paper is strongly influenced by extensive discussion with colleagues at Valence Labs and Recursion, as well as with current and former interns. In particular, we would like to thank Frederik Wenkel, Cian Eastwood, Lu Zhu, Cristian Gabellini, Hatem Helal, Julien Roy, Maciej Sypetkowski, Jessica Dafflon, Wilson Tu, Austin Tripp, Kerstin Klaeser, Craig Russell, Michel Moreau, Stephan Thaler, Malika Srivastava, Thomas Rochefort-Beaudoin, Yassir El Mesbahi, Ihab Bendidi, Honoré Hounwanou, Julien St-Laurent, Cassandra Masschelein, Sébastien Giguère, Soumali Roychowdhury, Véronique Bérubé, Francesco Di Giovanni, Therence Bois, Marta Fay for the extensive discussions. %

Finally, we are very grateful for the continued feedback and guidance provided by our scientific advisors Yoshua Bengio and Michael Bronstein, who helped initiate and shape many of the key ideas presented here.

\bibliography{refs}
\bibliographystyle{valence}

\newpage

\appendix

\section{Considerations for benchmarking virtual cells}\label{sec:benchmarking-considerations}

Here we present both scientific and practical considerations for benchmarking virtual cells. We begin by acknowledging that creating useful benchmarks in biology presents unique challenges not often found in other areas of ML/AI. In particular: 

\begin{itemize}
    \item Ground truth observations are often missing, incomplete, or biased
    \item Most biological observations are often static, capturing only a single point in time
    \item Many biological factors are highly dependent on each other, shaped by complex evolutionary pressures
    \item Biological data is often highly variable, due to a combination of both technical and systemic issues.
\end{itemize}

These factors make benchmarking virtual cells substantially more difficult than evaluating models in domains like vision or language, where data is more abundant, better characterized, and largely static. Moreover, we argue that they expose critical limitations in many existing benchmarks of virtual cell performance. For example, many of the datasets used in current benchmarks are biased toward transcriptomic readouts, primarily because they are the largest publicly-available resources. However, these datasets typically span only a narrow range of cellular contexts and perturbations. More fundamentally, transcriptomic changes alone capture only a partial view of cellular functional responses, as mRNA levels often do not correlate with protein activity, phenotypic outcomes, or dynamic state shifts, due to post-transcriptional regulation and other factors \citep{vogel2012insights,liu2016dependency,buccitelli2020mrnas}. Despite these limitations, the same transcriptomic datasets are reused across benchmarks without sufficient consideration of their biases, leading to correlated, easy-to-obtain metrics that do not meaningfully drive improvement toward predictive, explanatory, and actionable models.

We also note that it can be challenging to perform better than simple baselines\footnote{E.g., the average response of all perturbed training samples.}, and that the most performant models are not always the most advanced from an ML perspective, e.g., simple models like scVI \citep{lopez2018deep} already offer strong baselines when making predictions about transcriptomic data \citep{bendidi2024benchmarking}. We therefore strongly recommend that all virtual cell models begin their evaluations by comparing to simple baselines in order to assess whether they can sufficiently represent the data space.

In general, we believe that virtual cell benchmarks should be:

\begin{itemize}
    \item As generally-applicable as possible by adhering to the principles described in \Cref{sec:benchmarking-considerations-scientific} and \Cref{sec:benchmarking-considerations-practical}.
    \item As specifically-defined as possible by being associated with one or more of the capabilities described in \Cref{sec:benchmarking}.
    \item Reviewed and updated frequently in order to maintain relevance as biological knowledge and datasets evolve.
\end{itemize}

\subsection{Scientific considerations}\label{sec:benchmarking-considerations-scientific}

The following principles address scientific considerations for virtual cell benchmarks:

\paragraph{Focus on functional response.} Evaluation metrics should prioritize functional outputs that are directly relevant to drug efficacy.

\paragraph{Favor explainability.} Whenever possible, benchmarks should encourage explainable models, since explainability will drive biological understanding beyond the ability to predict.

\paragraph{Be biologically consistent.} A virtual cell should recapitulate widely-accepted biology. For example, models should generally respect known relationships between molecular layers (such as typical correlations between transcript level and protein abundance), but also accommodate cases where regulation occurs post-transcriptionally, translationally, or through protein degradation. Benchmarks should penalize grossly-implausible biological predictions without rigidly enforcing incomplete or oversimplified biological graphs.

\paragraph{Be statistically significant.} Conclusions should be statistically valid and supported by data. Evaluations should be sufficiently powered to declare statistical significance, and not encourage the comparison of mean performance alone (as is typically done using the common “bold table” format).

\paragraph{Be independent of modality.} To avoid modality-specific biases, benchmarks should be defined in a way that does not require a specific data modality to produce the measure, promoting models that generalize across different types of biological measurements.

\paragraph{Have biologically informed splits.} When possible, virtual cell benchmarks should design biologically-informed splits, e.g., use time-based splits to simulate knowledge acquisition, or evolution-based splits to account for the relationship between genes and their products.

\paragraph{Have complementary metrics.} When possible, evaluation should report multiple complementary metrics, as biological systems often behave in counterintuitive ways that are difficult to anticipate. A broad panel of metrics provides a more comprehensive view of model performance and robustness across diverse biological scenarios.

\paragraph{Have local and global metrics.} A good benchmarking suite should include both specific and high-level tasks to ensure that models are improving locally (e.g., predicting response of single-cell RNAseq readouts in HUVECs) and globally (e.g., recapitulating known biology across cell types).

\subsection{Practical considerations}\label{sec:benchmarking-considerations-practical}

The following principles address practical considerations for virtual cell benchmarks:

\paragraph{Update frequently.} The complexity of developing a benchmarking suite demands an iterative approach. With the help of interdisciplinary experts, the design of the entire benchmarking suite, including metrics, tasks, and data splits, needs to be periodically reviewed, refined, and updated based on factors like the desire for additional tasks and the availability of new datasets.

\paragraph{Encourage prospective evaluation.} Prospective evaluation on a blind test set is the gold standard for unbiased evaluation in ML research. Regularly-organized competitions have proven extremely useful in driving progress by providing opportunities to assess progress and disseminate results that inform future research directions. For example, CASP played a critical role in driving progress on protein structure prediction.

\paragraph{Simplify adoption.} Benchmarking suites should be made accessible via production-ready, open-source software, with modular, standardized components to encourage adoption, reproducibility, and transparency. Various software modules should provide a foundation for follow-up research and should make independently-generated results more comparable by standardizing the evaluation logic and operating as a source of truth.

\end{document}